\def\year{2021}\relax
\documentclass[letterpaper]{article} 
\usepackage{aaai21}  
\usepackage{times}  
\usepackage{helvet} 
\usepackage{courier}  
\usepackage[hyphens]{url}  
\usepackage{graphicx} 
\usepackage{natbib}  
\usepackage{caption} 
\usepackage{algorithm}
\usepackage{algorithmic}
\usepackage{amsmath}
\usepackage{amssymb}
\usepackage{booktabs} 
\usepackage{cleveref}
\usepackage{eqparbox}
\usepackage{latexsym}
\usepackage{microtype}
\usepackage{nicefrac}
\usepackage{placeins}
\usepackage{stmaryrd}
\usepackage{subfigure}
\usepackage{times}
\usepackage{xcolor}
\usepackage{xspace}

\DeclareMathOperator*{\argmin}{arg\,min}
\renewcommand{\algorithmiccomment}[1]{\hfill\eqparbox{COMMENT}{\it \# #1}}

\newcommand{\nmet}{Moving Targets\xspace}

\title{Teaching the Old Dog New Tricks: \\
Supervised Learning with Constraints}
\author {
        Fabrizio Detassis,\textsuperscript{\rm 1}
        Michele Lombardi, \textsuperscript{\rm 1}
        Michela Milano \textsuperscript{\rm 1} \\
}
\affiliations {
    \textsuperscript{\rm 1} DISI, University of Bologna \\
    fabrizio.detassis2@unibo.it, michele.lombardi2@unibo.it, michela.milano@unibo.it
}

\begin{document}

\maketitle

\begin{abstract}
Adding constraint support in Machine Learning has the potential to address outstanding issues in data-driven AI systems, such as safety and fairness. Existing approaches typically apply constrained optimization techniques to ML training, enforce constraint satisfaction by adjusting the model design, or use constraints to correct the output. Here, we investigate a different, complementary, strategy based on ``teaching'' constraint satisfaction to a supervised ML method via the direct use of a state-of-the-art constraint solver: this enables taking advantage of decades of research on constrained optimization with limited effort. In practice, we use a decomposition scheme alternating master steps (in charge of enforcing the constraints) and learner steps (where any supervised ML model and training algorithm can be employed). The process leads to approximate constraint satisfaction in general, and convergence properties are difficult to establish; despite this fact, we found empirically that even a na\"{i}ve setup of our approach performs well on ML tasks with fairness constraints, and on classical datasets with synthetic constraints.
\end{abstract}

\section{Introduction}%
\label{sec:Introduction}

Techniques to deal with constraints in Machine Learning (ML) have the potential to address outstanding issues in data-driven AI methods: they can boost generalization (e.g. if they represent physical laws), encode negative patterns (e.g. excluded classes) and relational information (e.g. involving multiple examples); they can ensure the satisfaction of desired properties, such as fairness, safety, or lawfulness.


To the best of the authors' knowledge, existing approaches for taking into account constraints in ML typically work by adapting ideas from constrained optimization to training algorithms/loss functions, or adjusting the model design, or by correcting the model output. Here we propose a different, complementary, strategy that enforces constraints in supervised ML by making direct use of any state-of-the-art constraint solver: this \emph{enables taking advantage of decades of research on constraint optimization} with limited effort.

Our method, referred to as \emph{\nmet}, is decomposition-based and alternates master and learner steps. The master step (addressed with the constraint solver) handles constraint satisfaction by adjusting the targets; the learner step trains a supervised ML model. Master and learner are isolated and communicate only via the vector of targets, so that: 1) any ML method can be used for the learner, with no modifications; 2) the master can rely on techniques such as Mathematical or Constraint Programming, which natively support complex constraints (including discrete and non-differentiable ones). Our method is also well suited to deal with relational constraints over large populations (e.g. fairness indicators). 


When constraints conflict with the data, the present approach \emph{prioritizes constraint satisfaction over accuracy}: for this reason, it is not well suited for exploiting fuzzy symbolic knowledge, unlike many approaches in the literature. Due to our open setting it is hard to determine convergence properties; despite this, we found that even a na\"{i}ve setup of the approach performs well (compared to state-of-the-art methods) on classification and regression tasks with fairness constraints, and on classification problems with balance constraints.

Due to its combination of simplicity, generality, and the observed empirical performance, \nmet can represent a valuable addition to the arsenal of techniques for dealing with constraints in Machine Learning.
The paper is organized as follows: in \Cref{sec:Related Works} we briefly survey related works on the integration of constraints in ML; in \Cref{sec:Method} we present our method and in \Cref{sec:Empirical Evaluation} our empirical evaluation. Concluding remarks are in \Cref{sec:Conclusion}.

\section{Related Works}%
\label{sec:Related Works}


Most approaches in the literature build on just a few key ideas. One of them is \emph{using the constraints to adjust the output of a trained ML model}. This is done in DeepProbLog~\cite{manhaeve2018deepproblog}, where Neural Networks with probabilistic output (mostly classifiers) are treated as predicates. \cite{rocktaschel2017end} presents a Neural Theorem Prover using differentiable predicates and the Prolog backward chaining algorithm. The original Markov Logic Networks \cite{richardson2006markov} rely instead on Markov Fields defined over First Order Logic formulas. As a drawback, with these approaches the constraints have no effect on the model parameters, which complicates the analysis of feature importance. Moreover, dealing with relational constraints (e.g. fairness) requires access at prediction time either to a representative population or to its distribution \cite{hardt2016equality, fish2016confidence}.

Other approaches operate by \emph{using constraint-based expressions as regularization terms during training}. In Semantic Based Regularization \cite{diligenti2017semantic} constraints are expressed as fuzzy logical formulas over differentiable predicates. Logic Tensor Networks \cite{serafini2016logic} focus on Neural Networks and replace the entire loss function with a fuzzy formula. Differentiable Reasoning \cite{vanKrieken2019semi} uses in a similar fashion relational background knowledge to benefit from unlabeled data. In the context of fairness constraints, this approach has been taken in \cite{aghaei2019learning, dwork2012fairness, zemel2013learning, calders2010three, kamiran2010discrimination}. These methods handle the constraints by adjusting the model parameters, and can therefore be used to analyze feature importance. They can deal with relational constraints without additional examples at prediction time; however, they require \emph{simultaneous} access at training time to large groups of examples linked by the constraints (which can be problematic when using mini-batches). They often require properties on the constraints (e.g. differentiability), which may force approximations; they may also be susceptible to numerical issues.

A third idea consists in \emph{enforcing constraint satisfaction in the data via pre-processing}. This is proposed in the context of fairness constraints by \cite{kamiran2009classifying, DBLP:journals/kais/KamiranC11, luong2011k}. The approach enables the use of standard ML methods with no modification, and can deal with relational constraints on large sets of examples. As a main drawback, the model/training algorithm may have trouble approximating the revised labels, leading to substantial degrees of infeasibility.


\emph{Multiple ideas can be combined}: domain knowledge has been introduced in differentiable Machine Learning (e.g. Deep Networks) by designing their structure, rather than the loss function: examples include Deep Structured Models in \cite{lin2016efficient} and \cite{Xuezhe2016}. These approaches can use constraints to support both training and inference.



\begin{table}[tb]
\begin{center}
\begin{small}
\begin{tabular}{ccc}
\toprule
\bf Loss Function & \bf Expression & \bf Target Space \\
\midrule
Mean Squared Error & $\displaystyle \frac{1}{m} \|y - y^*\|_2^2$ & $\mathbb{R}^m$ \\
Hamming Distance & $\displaystyle \frac{1}{m} \sum_{i=1}^m {\rm I}[y_i \neq y^*_j] $ & $\{1..c\}^m$ \\
Cross Entropy & $\displaystyle \frac{1}{m} \sum_{i=1}^m \sum_{j=1}^ c y^*_{ij} \log y_{ij} $ & $[0, 1]^m$ \\
\bottomrule
\end{tabular}
\end{small}
\end{center}
\caption{Notable losses ($m=\text{\# examples}$, $c = \text{\#classes}$)}
\label{tab:losses}
\end{table}

\section{\nmet}%
\label{sec:Method}

In this section we present our method, discuss its properties and provide some convergence considerations.

\paragraph{The Algorithm}

Our goal is to adjust the parameters of a ML model so as to minimize a loss function for supervised learning, under a set of generic constraints. We acknowledge that any constrained learning problem must trade prediction mistakes for a better level of constraint satisfaction, and we attempt to \emph{control this process by carefully selecting which mistakes should be made}. This is similar in spirit to \cite{kamiran2009classifying, DBLP:journals/kais/KamiranC11, luong2011k}, but: 1) we consider generic constraints rather than focusing on fairness; 2) we consider generic supervised learning rather than just binary classification; 3) we rely on an iterative process (which alternates ``master'' and ``learner'' steps) to improve the results.


Let $L(y, y^*)$ be the loss function, where $y$ is the prediction vector and $y^*$ is the target vector. We make the (non-restrictive) assumption that \emph{the loss is a pre-metric} -- i.e. $L(y, y^*) \geq 0$ and $L(y, y^*) = 0$ iff $y = y^*$. Examples of how to treat common loss functions can be found in \Cref{tab:losses}.

We then want to solve, in an exact or approximate fashion, the following constrained optimization problem:
\begin{equation}
   \argmin_{\theta} \{L(y, y^*) \mid y = f(X, \theta), y \in C\} \label{eq:start}
\end{equation}
where $f$ is the ML model and $\theta$ its parameter vector. With some abuse of notation we refer to $f(X, \theta)$ as the vector of predictions for the examples in the training set $X$. Since the model input at training time is known, constraints on both the model input and output can be represented as a feasible set $C$ for the sole predictions $y$.

The problem can be rewritten \emph{in pure target space}, without loss of generality, by introducing a second set $B = \{y \mid \exists \theta, y = f(X, \theta)\}$ corresponding to the ML model bias:
\begin{equation}
   \argmin_{y} \{L(y, y^*) \mid y \in B \cap C\} \label{eq:bap}
\end{equation}

\begin{algorithm}[tb]
\caption{{\sc\nmet}}
\begin{algorithmic}
   \INPUT label vector $y^*$, scalar parameters $\alpha, \beta, n$
   \STATE $y^1 = l(y^*)$ \COMMENT{pretraining}
   \FOR{$ k = 1..n $}
   \IF{$y^{k} \notin C$}
   \STATE $z^k = m_\alpha(y^k)$ \COMMENT{infeasible master step}
   \ELSE
   \STATE $z^k = m_\beta(y^k)$ \COMMENT{feasible master step}
   \ENDIF
   \STATE $y^{k+1} = l(z^k)$ \COMMENT{learner step}
   \ENDFOR
\end{algorithmic}
\label{alg:algo}
\end{algorithm}

The \nmet method is described in \Cref{alg:algo}, and starts with a learner step w.r.t. the original target vector $y^*$ (pretraining). Each learner step, given a target vector as input, solves approximately or exactly the problem:
\begin{equation}
   \mathit{l}(z) = \argmin_{y} \{L(y, z) \mid y \in B\} \label{eq:learner}
\end{equation}
Note that this is \emph{a traditional unconstrained learning problem}, since $B$ is just the model/algorithm bias. The result of the first learner step gives an initial vector of predictions $y^1$.

Next comes a master step to adjust the target vector: this can take two forms, depending on the current predictions. \emph{In case of an infeasibility}, i.e. $y^k \notin C$, we solve:
\begin{equation}
   \mathit{m_\alpha}(y) = \argmin_z \left\{L(z, y^*) + \frac{1}{\alpha} L(z, y) \mid z \in C \right\}
\end{equation}
I.e., we try to find a feasible label vector $z$ that balances the distance (in terms of loss) to both the original labels $y^*$ and the current prediction $y$. A parameter $\alpha \in (0, \infty)$ controls the trade-off. \emph{If the input vector is feasible} we instead solve:
\begin{equation}
   \mathit{m_\beta}(y) = \argmin_z \left\{L(z, y^*) \mid L(z, y) \leq \beta, z \in C \right\} \label{eq:master2}
\end{equation}
i.e. we look for a feasible label vector $z$ that is 1) not too far from the current predictions (in the ball defined by $L(z, y) \leq \beta$) and 2) closer (in terms of loss) to the true labels $y^*$. The differences from $\mathit{m_\alpha}(y)$ are needed to handle some corner cases (e.g. classification with accuracy loss).

\begin{figure}[tb]
\begin{center}
\centerline{\includegraphics[width=0.7\columnwidth]{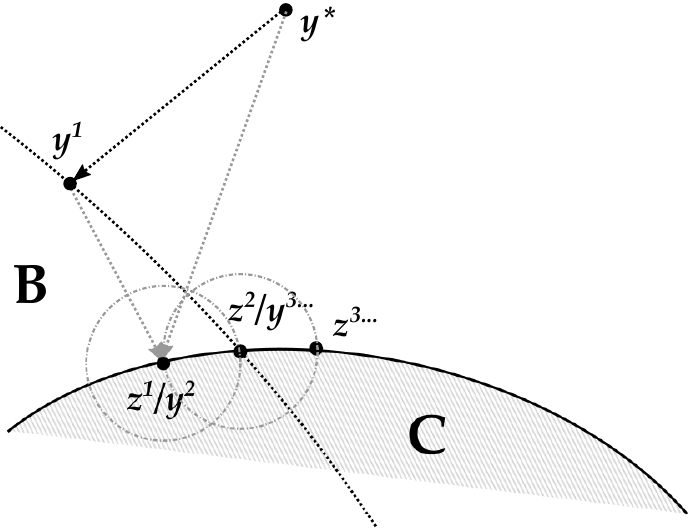}}
\caption{A sample run of our algorithm}
\label{fig:example}
\end{center}
\end{figure}

We then make a learner step trying to reach the adjusted labels; the new predictions will be adjusted at the next iteration and so on. In case of convergence, the predictions $y^k$ and the adjusted labels $z^k$ become stationary (but not necessarily identical). An example run, for a Mean Squared Error loss and convex constraints and bias, is in \Cref{fig:example}.

\paragraph{Discussion}

The learner problem is unconstrained, thus enabling the use of arbitrary ML approaches. The master problems do not need to deal with the ML model, making them far easier to solve for constrained optimization approaches. Since we make no explicit use of mini-batches, we can deal well with relational constraints on large groups (e.g. fairness). The master step can be addressed via any suitable solver, so that discrete variables and non-differentiable constraints can be tackled via (e.g.) Mathematical Programming, Constraint Programming, or SAT Modulo Theories.

Due to the very open setting, \emph{convergence properties are difficult to establish}. \Cref{eq:bap} is the Best Approximation Problem, while the learner step in \Cref{eq:learner} is a projection problem: this relates \nmet to the Alternating Projections (AP) method, Douglas-Rachford splits -- see e.g. \cite{boyd2003alternating} --, or the algorithm from \cite{artacho2018new}. Unfortunately, none of these approaches can be used directly, unless we introduce strong assumptions (e.g. convexity, lack of discrete predictions). Both forms of the master step are loosely derived from the Proximal Gradient Method \cite{parikh2014proximal}, and under restrictive assumptions should inherit its convergence properties. In practice, however, we are mostly concerned with non-convex ML models and complex constraints, meaning that at least the learner problem will be solved to local optimality. This limits our interest in a formal convergence analysis.

Constraint satisfaction guarantees cannot be provided in general, since the intersection $B \cap C$ in \Cref{eq:bap} could be empty. Even if that is not the case, as a side effect of using a decomposition and relying (in most practical cases) to a non-exact learner, our method may fail to reach constraint satisfaction. In practice, \nmet usually reaches feasibility or near-feasibility in our empirical evaluation.

Depending on the constraints, loss, and the target space the master problems may be NP-hard. Even in this case, state-of-the-art solvers may find exact solutions for datasets of practical size. Moreover, for separable loss functions (e.g. all those from \Cref{tab:losses}), the master problems can be defined over only the constrained examples, with a possibly significant size reduction. If scalability is still a concern, the master step can be solved to near-optimality via heuristics, meta-heuristics or truncated exact algorithms. Given that the learner problem is also likely solved to local optimality, using non-exact methods in the master is not in principle a critical concern.

\section{Empirical Evaluation}%
\label{sec:Empirical Evaluation}

Here we describe our experimentation, which is designed around a few main questions: 1) How does the method work on a variety of constraints, tasks, and datasets? 2) What is the effect of the $\alpha, \beta$ parameters? 3) How does the approach scale? 4) How different is the behavior with different ML models? 5) How does the method compare with alternative approaches? Our code and results are publicly available\footnote{Git repository available at github.com/fabdet/moving-targets}.

\paragraph{Tasks and Constraints}%
We experiment on three case studies. First, we consider a (synthetic) \emph{classification problem augmented with a balance constraint}, which forces the distribution over the classes to be approximately uniform. The loss function is the Hamming distance (accuracy) and the target space is $\{1..c\}^m$.  The $m_\alpha(y)$ problem is formulated as a Mixed Integer Linear Program (MILP) with binary variables $z_{ij}$ such that $z_{ij} = 1$ iff the adjusted class for the $i$-th example is $j$. Formally:
\begin{align}
    \min \ & \frac{1}{m} \sum_{i=1}^m (1 - z_{i,y^*_i}) + \frac{1}{\alpha m} \sum_{i=1}^m (1 - z_{i,y_i})
    \label{eq:ma1_obj} \\
    \text{s.t. } & \sum_{j=1}^c z_{ij} = 1 \hspace{22mm} \forall i = 1..m 
    \label{eq:ma1_mutex} \\
     & \sum_{i=1}^{m} y_{ij} \leq \left\lceil \frac{(1 + \xi)m}{c} \right\rceil \hspace{10mm} \forall j=1..c 
     \label{eq:ma1_bal} \\
     & z_{ij} \in \{0, 1\} \hspace{8.5mm} \forall i=1..m, j=1..c
\end{align}
The summations in \Cref{eq:ma1_obj} encode the Hamming distance w.r.t. the true labels $y^*$ and the predictions $y$. \Cref{eq:ma1_mutex} prevents assigning two classes to the same example. \Cref{eq:ma1_bal} requires an equal count for each class, with tolerance defined by $\xi$ ($\xi = 0.05$ in all our experiments);
the balance constraint is stated in exact form, thanks to the discrete labels. The $m_\alpha$ formulation generalizes the knapsack problem and is hence NP-hard; since all examples appear in \Cref{eq:ma1_bal}, no problem size reduction is possible. The $m_\beta$ problem can be derived from $m_\alpha$ by changing the objective function and by adding the ball constraint as in \Cref{eq:master2}.

Our second use case is \emph{a classification problem with realistic fairness constraints}, based on the DIDI indicator from \cite{aghaei2019learning}:
\begin{align}
    & \mathit{DIDI}^c(X, y) =  \sum_{k \in K} \sum_{v \in D_k} \sum_{j = 1}^c d_{kvj} \label{eq:didic} \\
    & d_{k,v,j} = 
    \left| \frac{1}{m} \sum_{i = 1}^m {\rm I}[y_i = j] -
    \frac{1}{|X_{k,v}|} \sum_{i \in X_{k,v}} {\rm I}[y_i = j]\ \right| \nonumber
\end{align}
where $K$ contains the indices of ``protected features'' (e.g. ethnicity, gender, etc.). $D_k$ is the set of possible values for the $k$-th feature, and $X_{k,v}$ is the set of examples having value $v$ for the $k$-th feature. 
The DIDI indicator measures whether there exists a disparate outcome for examples belonging to protected groups; this gap is null for unbiased models.
The $m_\alpha(y)$ problem can be defined via the following Mathematical Program:
\begin{align}
    \min \ & \frac{1}{m} \sum_{i=1}^m (1 - z_{i,y^*_i}) + \frac{1}{\alpha m} \sum_{i=1}^m (1 - z_{i,y_i}) \\
    \text{s.t. } & \text{\Cref{eq:ma1_mutex}} \nonumber \\
     & \sum_{k \in K} \sum_{v \in D_k} \sum_{j=1}^c d_{kvj} \leq \epsilon \hspace{10mm} \forall j=1..c \label{eq:ma2_didi} \\
     & d_{kvj} = \left| \sum_{i = 1}^m \frac{y_{ij}}{m} -
     \sum_{i \in X_{k,v}} \frac{y_{ij}}{|X_{k,v}|} \right|
     \label{eq:ma2_abs1} \\
     & z_{ij} \in \{0, 1\} \hspace{8.5mm} \forall i=1..m, j=1..c
\end{align}
where \Cref{eq:ma2_didi} is the constraint on the DIDI value and \Cref{eq:ma2_abs1} is then linearized using standard MILP methods.
The DIDI scales with the number of examples and has an intrinsic value due to the discrimination in the data.
Therefore, we compute $\mathit{DIDI}_{tr}$ for the training set, then in our experiments we have $\epsilon = 0.2 \mathit{DIDI}_{tr}$. This is again an NP-hard problem defined over all training examples. The $m_\beta$ formulation can be derived as in the previous case.

Our third case study is \emph{a regression problem with fairness constraints}, based on a specialized DIDI version from \cite{aghaei2019learning}:
\begin{align}
    & \mathit{DIDI}^r(X, y) =  \sum_{k \in K} \sum_{v \in D_k} d_{kv} \\
    & d_{k,v,j} = 
    \left| \frac{1}{m} \sum_{i = 1}^m y_i -
    \frac{1}{|X_{k,v}|} \sum_{i \in X_{k,v}} y \ \right|
    \label{eq:didir}
\end{align}
In this case, we use the Mean Squared Error (MSE) as a loss function, and the label space is $\mathbb{R}^m$. The $m_\alpha$ problem can be defined via the following Mathematical Program:
\begin{align}
    \min \ & \frac{1}{m} \sum_{i=1}^m (y^*_i - z_{i})^2 + \frac{1}{\alpha m} \sum_{i=1}^m (z_{i} - y_i)^2 \\
    \text{s.t. } & \sum_{k \in K} \sum_{v \in D_k} d_{kv} \leq \epsilon \hspace{10mm} \forall j=1..c \label{eq:ma3_didi} \\
     & d_{kv} = \left| \sum_{i = 1}^m \frac{y_{i}}{m} -
     \sum_{i \in X_{k,v}} \frac{y_{i}}{|X_{k,v}|} \right|
     \label{eq:ma3_abs1} \\
     & z_{i} \in \mathbb{R} \hspace{8.5mm} \forall i=1..m
\end{align}
After a standard reformulation of \Cref{eq:ma3_abs1}, this is a linearly constrained, convex, Quadratic Programming problem that can be solved in polynomial time. The $m_\beta$ problem can be derived as in the previous cases: while still convex, $m_\beta$ is in this case a Quadratically Constrained Problem.

\begin{table*}[tb]
\centering
\resizebox{\textwidth}{!}{
\begin{tabular}{lr|ccccccc}
\toprule
NN ($\alpha$, $\beta$)  
& & Ptr & $\alpha=1$ & $\alpha=1$  & $\alpha=1$ & $\alpha=.1$  & $\alpha=0^+$ & Ideal case  \\
& & & $\beta = .01$ & $\beta = .05$ & $\beta = .1$ & $\beta = .01$ & $\beta = 0.1 $ \\
\midrule
Iris
&S
& $.970 \pm .002$ & $.99 \pm .01$ & $\mathbf{.997} \pm .004$ & $\mathbf{.997} \pm .004$ & $.99 \pm .02$ & $0.995 \pm 0.008$ & $.9968 \pm .0004$\\
&C
& $.23 \pm .08$ & $.08 \pm .3$ & $.0 \pm .3$ & $.0 \pm .3$ & $.15 \pm .4$ &  $.0 \pm .3$ & $.0 \pm .3$ \\
\midrule
Redwine
&S
& $.709 \pm .005$ & $.508 \pm .006$ & $\mathbf{.511} \pm .009$ & $.506 \pm .006$ & $.484 \pm .007$  & $.50 \pm .01$ & $.525 \pm .002$\\
&C
& $.05 \pm .05$ & $.0 \pm .05$ & $.0 \pm .03$ & $.0 \pm .04$ & $.0 \pm .02$ & $.0 \pm .05$ & $.0 \pm 0$ \\
\midrule
Whitewine 
&S
& $.644 \pm .002$ & $\mathbf{.446} \pm .006$ & $.437 \pm .009$ & $.439 \pm .009$ & $.40 \pm .02$ & $.401 \pm .009$ & $.524 \pm .002$\\
&C
& $1^{+} \pm .2$ & $.0 \pm .1$ & $.0 \pm .3$ & $.0 \pm .2$ & $.0 \pm .3$ & $.0 \pm .3$ & $.0 \pm .1$\\
\midrule
Shuttle
&S
& $.999 \pm 0$ & $.39 \pm .04$ & $.37 \pm .01$ & $.375 \pm .007$ & $.37 \pm .03$ & $.37 \pm .03$ & $.3608 \pm .0008$\\
&C
& $1^{+} \pm 0$ & $1^{+} \pm 1$ & $.7 \pm .2$ & $.6 \pm .4$ & $1^{+} \pm 1$ & $1^{+} \pm 1$ & $0 \pm 0$\\
\midrule
Dota2
&S
& $.686 \pm .002$ & $.666 \pm .007$ & $.661 \pm .002$ & $.66 \pm .01$ & $\mathbf{.672} \pm .004$ & $.656 \pm .006$ & $.9984 \pm .0009$\\
&C
& $1^{+} \pm .3$ & $.6 \pm 1$ & $.6 \pm 1$ & $1^{+} \pm 1$ & $.0 \pm .2$ & $1^{+} \pm 1$ & $.0 \pm 0$\\
\midrule
Adult
&S
& $.867 \pm 0.001$ & $.818 \pm .005$ & $\mathbf{.86} \pm .02$ & $.841 \pm .006$ & $.852 \pm .004$ & $.84 \pm .02$ & $0.992 \pm .0005$\\
&C
& $1^{+} \pm .2$ & $.0 \pm .2$ & $.0 \pm .1$ & $.1 \pm .4$ & $.1 \pm .2$ & $.1 \pm .2$ & $0. \pm 0$\\
\midrule
Crime
&S
& $.56 \pm .02$ & $\mathbf{.49} \pm .01$ & $.46 \pm .04$ & $.48 \pm .03$ & $.45 \pm .05$ & $.46 \pm .06$ & $.910 \pm .007$\\
&C
& $1^{+} \pm .1$ & $.1 \pm .4$ & $.0 \pm .4$ & $.0 \pm .5$ & $.0^ \pm .1$ & $.05 \pm .2$ & $.0 \pm 0$\\
\bottomrule
\end{tabular}
}
\caption{Effect of parameters $\alpha$ and $\beta$ on different datasets.}
\label{tab:parameter_tuning}
\end{table*}

\paragraph{Datasets, Preparation, and General Setup}

We test our method on seven datasets from the UCI Machine Learning repository \cite{Dua:2019}, namely \textit{iris} (150 examples), \textit{redwine} (1,599), \textit{crime} (2,215), \textit{whitewine} (4,898), \textit{adult} (32,561), \textit{shuttle} (43,500), and \textit{dota2} (92,650). We use \emph{adult} for the classification/fairness case study, \emph{crime} for regression/fairness, and the remaining datasets for the classification/balance case study.

For each experiment, we perform a 5-fold cross validation (with a fixed seed). Hence, the training set for each fold will include $80\%$ of the data. All our experiments are run on an Intel Core i7 laptop with 16GB RAM and no GPU acceleration, and we use Cplex 12.8 to solve the master problems. For sake of simplicity, we opted for straightforward setup of the constraint solver (default parameters, exact solution of even NP-hard problems).


All the datasets for the classification/balance case study are prepared by standardizing all input features (on the training folds) to have zero mean and unit variance. The \emph{iris} and \emph{dota2} datasets are very balanced, while the remaining datasets are quite unbalanced.
In the \emph{adult} (also known as ``Census Income'') dataset the target is ``income'' and the protected attribute is ``race''. We remove the features ``education'' (duplicated) and ``native country'' and use a one-hot encoding on all categorical features.
Features are normalized between 0 and 1.
Our \emph{crime} dataset is the ``Communities and Crime Unnormalized'' table. The target is ``violentPerPop'' and the protected feature is ``race''.
We remove features that are empty almost everywhere and features trivially related to the target (``murders'', ``robberies'', etc.). Features are normalized between 0 and 1 and we select the top 15 ones according to the \texttt{SelectKBest} method of scikit-learn (excluding ``race''). The protected feature is then reintroduced.

\paragraph{Parameter tuning}

We perform an investigation of the impact of $\alpha$ and $\beta$ by running the algorithm for 15 iterations (used in all experiments), with different parameter values.
As a ML model, we use a fully-connected, feed-forward Neural Network (NN) with two hidden layers with 32-Rectifier Linear Units. 
The last layer has either a SoftMax activation (for classification) or Linear (for regression). The loss function is respectively the categorical cross-entropy or the MSE.
The network is trained with 100 epochs of RMSProp in Keras/Tensorflow 2.0 (default parameters, batch size 64). 

The results are in \Cref{tab:parameter_tuning}. We report a score (row \emph{S}, higher is better) and a level of constraint violation (row \emph{C}, lower is better). The \emph{S} score is the accuracy for classification and the R2 coefficient for regression. For the balance constraint, the \emph{C} score is the standard deviation of the class frequencies; in the fairness case studies, we use the ratio between the DIDI of the predictions and that of the training data. Both indicators are then normalized over the constraint satisfaction threshold, and capped at 1 for readability (capped values are marked as $1^+$). Cells report mean and standard deviation for the 5 runs.



All columns labeled with $\alpha$ and $\beta$ values refer to our method with the specified parameters. The \emph{ideal case} refers to a simple projection of the true target $y^*$ on the feasible space $C$. This corresponds to an upper bound on the performance of a constrained learner: it exactly matches the constraint threshold while minimizing the loss function.  The \emph{ptr} column reports the results of the pretraining step, as defined in \cref{alg:algo}, i.e. a constraint-agnostic behavior. Our method lies inbetween the two extreme cases. Accuracy comparisons are fair only for similar constraint violation scores.

The \nmet algorithm can \emph{significantly improve the satisfaction of non-trivial constraints}: this is evident for the unbalanced datasets \emph{redwine}, \emph{whitewine}, and \emph{shuttle} and all fairness use cases, for which feasible (or close) results are almost always obtained. As one can expect, satisfying very tight constraints (e.g. in the unbalanced dataset) comes at a steep cost in terms of accuracy. 
Finally, \emph{reasonable parameter choices have only a mild effect on the algorithm behavior}, thus simplifying its configuration. Empirically, $\alpha = 1, \beta = 0.1$ seems to works well and is used for all subsequent experiments.

\begin{figure}[bt]
    \centering
    \includegraphics[width=0.75\columnwidth]{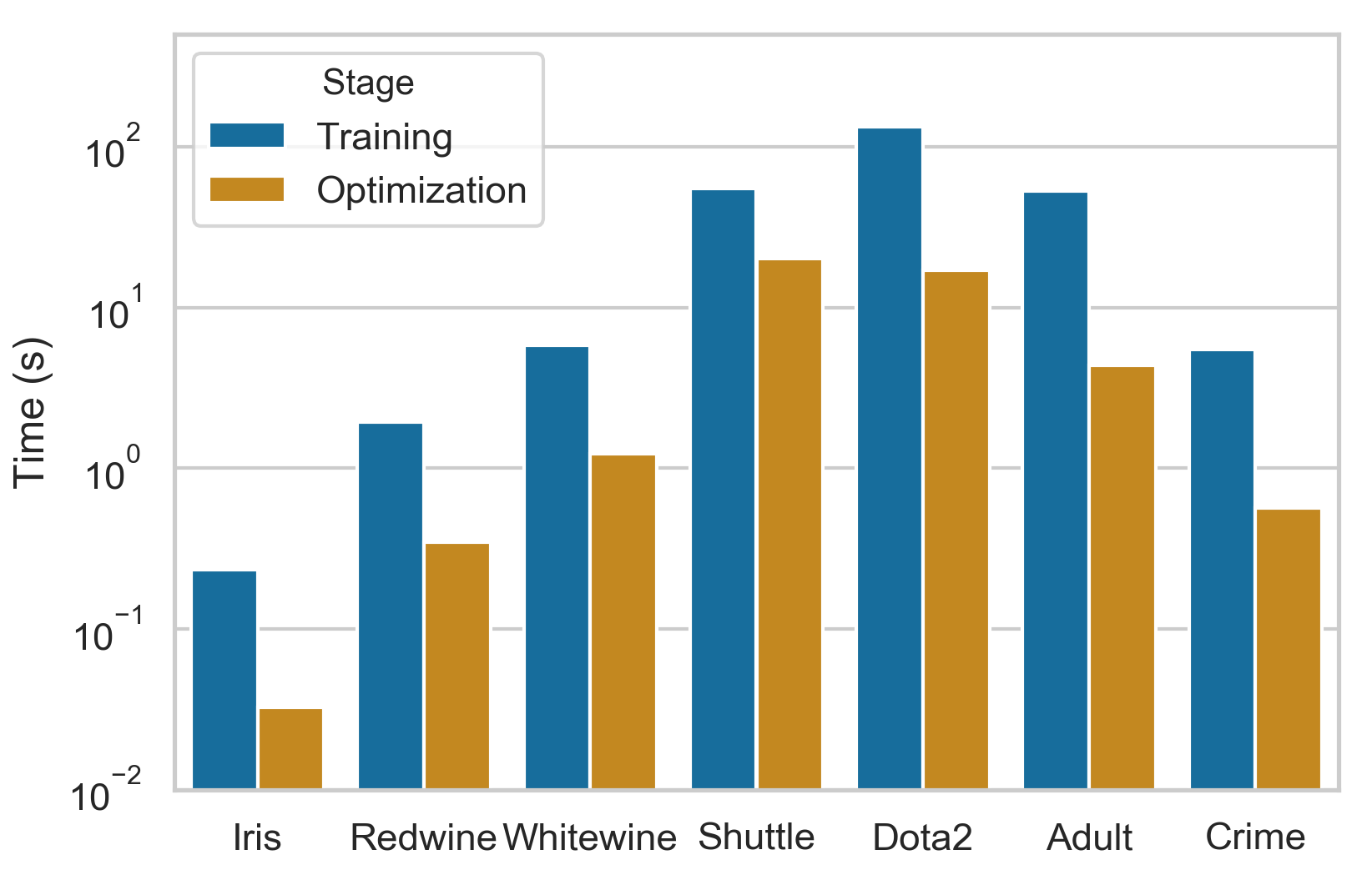}
    \caption{Average master step time, compared to NN training}
    \label{fig:computational_time}
\end{figure}

\paragraph{Scalability}

We next turn to investigating the method scalability. Our examples can be considered worst cases, since all examples appear in the single constraints and in some case involve NP-hard problems. We report the average time for a master step in \Cref{fig:computational_time}, with average time for a learner step (100 epochs of our NN) for reference.
At least in our experimentation, \emph{the time for a master step is always very reasonable}, even for the \emph{dota2} dataset for which we solve NP-hard problems on 74,120 examples. This is mostly due to the clean structure of the $m_\alpha$ and $m_\beta$ problems. Of course, for sufficiently large training sets, exact solutions will become impractical and non-exact optimization will need to be considered (e.g. meta-heuristics or matheuristics).

\paragraph{Setup of Alternative Approaches}

Here we describe the setup of alternative approaches that will be used for comparison. Namely, we consider the regularized linear approach from \cite{DBLP:journals/corr/BerkHJJKMNR17}, referred to as RLR, a Neural Network with Semantic Based Regularization \cite{diligenti2017-01}, referred to as SBR, and the Lagrangian approach from \cite{cotter2019optimization}, referred to as TFCO. The first two approaches introduce constraints as regularizers at training time. Their loss function is in the form:
\begin{equation}
    L(f(X; \theta), y^*) + \mu g(f(X; \theta))
    \label{eq:regularizer}
\end{equation}
The regularization term must be differentiable and the multiplier $\mu$ needs to be hand-tuned. The TFCO approach is similar, but it optimizes both the model parameters and the multipliers by alternating loss minimization and constraint satisfaction.

We use SBR only for the case studies with the balance constraint, which we are forced to approximate to obtain differentiability:
\begin{equation}
    g(f(X; \theta)) = \max_{j = 1..c} \sum_{i = 1}^m f(X; \theta)
\end{equation}
i.e. we use the sums of the NN output neurons to approximate the class counts and the maximum as a penalty; this proved superior to other attempts in preliminary tests. The $L$ term is the categorical cross-entropy.

\begin{table}[tb]
\centering
\begin{tabular}{r|lr|ccc}
\toprule
&& $\mu$ & $0.01$ & $0.1$ & $1$ \\
\midrule
SBR 
&
Iris 
& S & $\mathbf{0.984}$ & $0.97$ & $0.4$ \\
&& C & $0$ & $1$ & $1^{+}$\\
&Redwine 
& S & $0.15$ & $0.15$ &  $\mathbf{0.17}$\\
&& C & $1^{+}$ & $1^{+}$ & $1$\\
&Whitewine 
& S & $0.17$ & $\mathbf{0.15}$ & $0.14$\\
&& C & $1^{+}$ & $0.3$ & $1$\\
&Shuttle 
& S & $0.7$ & $\mathbf{0.31}$ & $0.14$\\
&& C & $1^{+}$ & $0.8$ & $0.8$\\
&Dota2 
& S & $\mathbf{0.61}$ & $0.48$ & $0.49$\\
&& C & $1^{+}$ & $1^{+}$ & $1^{+}$\\
\midrule
RLR
&
Adult 
& S & $\mathbf{.83} $ & $.75$ & $.75$\\
&& C & $1^{+}$ & $1^{+}$ & $1^{+}$\\
&Crime 
& S & $.39$ & $\mathbf{0.30}$ & $\mathbf{0.30}$\\
&& C & $1$ & $0$ & $0$\\
\bottomrule
\end{tabular}
\caption{Effect of parameter $\mu$ in regularization methods.}
\label{tab:regtuning}
\end{table}

Our SBR approach relies on the NN model from the previous paragraphs. Since access to the network structure is needed to differentiate the regularizer, SBR works best when all the examples linked by relational constraints can be included in the same batch. When this is not viable the regularizer can be treated stochastically (via subsets of examples), at the cost of additional approximation. We use a batch size of 2,048 as a compromise between memory usage and noise. The SBR method is trained for 1,600 epochs.

The RLR approach relies on linear models (Logistic or Linear Regression), which are simple enough to consider large group of examples simultaneously. We use this approach for the fairness use cases. In the \emph{crime} (regression) dataset $L$ is the MSE and the regularizer is simply \Cref{eq:didir}. In the \emph{adult} (classification) dataset $L$ is the cross-entropy; the regularizer is \Cref{eq:didic}, with the following substitution:
\begin{align*}
    & d_{k,v,j} = 
    \left| \frac{1}{m} \sum_{i = 1}^m \theta^\top x_i -
    \frac{1}{|X_{k,v}|} \sum_{i \in X_{k,v}} \theta^\top x_i \right|
\end{align*}
This is an approximation obtained according to \cite{DBLP:journals/corr/BerkHJJKMNR17} by disregarding the sigmoid in the Logistic Regressor to preserve convexity. We train this approach to convergence using the CVXPY 1.1 library (with default configuration).
In RLR and SBR classification, the introduced approximations \emph{permit to satisfy the constraints by having equal output for all classes}, i.e. completely random predictions. This undesirable behavior is countered by the $L$ term.


The results of a hand-tuning process for SBR and RLR are reported in \Cref{tab:regtuning}. In most cases, larger $\mu$ values tend as expected to result in better constraint satisfaction, with a few notable exceptions for classification tasks (\emph{iris}, \emph{dota}, and \emph{adult}). The issue is \emph{likely due to the approximations introduced in the regularizers}, since it arises even on small datasets that fit in a single mini-batch (\emph{iris}). Further analysis will be needed to confirm this intuition. The accuracy decreases for a larger $\mu$, as expected, but at a rather rapid pace. In the subsequent experiments, \emph{we will use for each dataset the RLR and SBR that offer the best accuracy while being as close to feasible as possible}: these are the cells in bold font in \Cref{tab:regtuning}.
For the TFCO approach, we use again the NN from previous paragraphs, a minibatch of size 200 and 100 iterations with 200 iterations per loop. The optimizer is ADAM with default parameters. The method is in principle able to reach an optimal solution, but \emph{only in expectation}, at the price of having a stochastic classifier. To enable a fair comparison, we obtain a single classifier using the ``best'' method from the reference implementation.

\begin{table*}[tb]
\centering
\begin{tabular}{lr|c|c|cccc}
\toprule
& & Regularized methods & TFCO & NN & LR & Ensemble trees & NN$_{pp}$ \\
\midrule
Iris
& S
& $.984 \pm .006$ & $.95 \pm .003$ & $\mathbf{.997} \pm .004$ & $.96 \pm .02$ & $.995 \pm .004$ & $.96 \pm .01$\\
& C
& $.0 \pm 0.2$ & $1^{+} \pm 1$ & $.0 \pm 0.3$ & $.1 \pm .4$ & $.0 \pm .2$ & $.07 \pm .4$\\
Redwine
& S
& $.17 \pm .05$ & $.3 \pm .2$ & $\mathbf{.506} \pm .006$ & $.32 \pm .01$ & $.40 \pm .02$ & $.480 \pm .001$\\
& C
& $.1^{+} \pm .5$ & $1^{+} \pm 1$ & $.0 \pm .05$ & $.6 \pm .2$ & $1^{+} \pm .5$ & $1^{+} \pm .3$\\
Whitewine
& S
& $.15 \pm .03$ & $.3 \pm .1$ & $\mathbf{.439} \pm .009$ & $.025 \pm .009$ & $.37 \pm .04$ & $.47 \pm .02$\\
& C
& $.3 \pm .3$ & $1^{+} \pm 0$ & $.0 \pm .2$ & $.8 \pm .2$ & $ 1^{+} \pm 1$ & $1^{+} \pm 1$\\
Shuttle
& S
& $.31 \pm .04$ & $.2 \pm .3$ & $.375 \pm .007$ & $\mathbf{.332} \pm .007$ & $ .51 \pm .05 $ & $.5 \pm .1$\\
& C
& $1 \pm 1$ & $1^{+} \pm 0$ & $.6 \pm .3$ & $ .4 \pm .4 $ & $1^{+} \pm .6$ & $1^{+} \pm 1$\\
Dota2 
& S
& $.61 \pm .02$ & $.53 \pm .01 $ & $.66 \pm .01$ & $.592 \pm .005$ & $ .53 \pm .01 $ & $\mathbf{.689} \pm .003$\\
& C
& $1^{+} \pm 1$ & $1^{+} \pm 0$ & $1^{+} \pm 1$ & $.5 \pm 0$ & $ 1^{+} \pm 1 $ & $.0 \pm .8$\\
Adult
& S
& $.834 \pm .001$ & $.87 \pm .01$ & $.841 \pm .006$ & $.805 \pm .006$ & $ .76 \pm .01 $ & $\mathbf{.865} \pm .003$\\
& C
& $1^{+} \pm .2$ & $1^{+} \pm .05$ & $.1 \pm .4$ & $.0 \pm .2$ & $ .0 \pm .2 $ & $.0 \pm .4$\\
Crime 
& S
& $.30 \pm .01$ & $\mathbf{.58} \pm .05$ & $.48 \pm .03$ & $.369 \pm .008$ & $.49 \pm .01$ & $.484 \pm .008$\\
& C
& $0 \pm 0$ & $.0 \pm .1$ & $.0 \pm .5 $ & $.0 \pm 0$ & $.2 \pm .05$ & $.0 \pm .1$\\
\bottomrule
\end{tabular}
\caption{Benchmarks with different ML models and alternative approaches}
\label{tab:benchmarks}
\end{table*}

\paragraph{Alternative Approaches and ML Models}

We can now compare the performance of \nmet using different ML models with that of the alternative approaches presented above, plus a pre-processing approach adapted from \cite{kamiran2009classifying}, referred to as $\rm NN_{pp}$ and obtained by setting $\alpha, \beta \rightarrow \infty$ in our method.


For our method, we consider the following ML models: 1) the NN from the previous section with $\alpha = 1, \beta = 0.1$; 2a) a Random Forest Classifier with 50 estimators and maximum depth of 5 (used for all classification case studies); 2b) a Gradient Boosted Trees model, with 50 estimators, maximum depth 4, and a minimum threshold of samples per leaf of 5 (for the regression case study); 4a) a Logistic Regression model (for classification); 4b) a Linear Regression model (for regression). All models except the NN are implemented using scikit-learn \cite{scikit-learn}. In \Cref{tab:benchmarks}, the tree ensemble method are reported on a single column, while another column (LR) groups Logistic and Linear regression.

Our algorithm seems to work well with all the considered ML models: tree ensembles and the NN have generally better constraint satisfaction (and higher accuracy for constraint satisfaction) than linear models, thanks to their larger variance. The preprocessing approach is effective when constraints are easy to satisfy (\emph{iris} and \emph{dota2}) and on all the fairness case studies, though less so on the remaining datasets. All \nmet approaches tend to perform better and more reliably than RLR and SBR.
The case of RLR and LR is particular, since in principle the two approaches can be expected to behave identically (convex problem  and same constraint formulation): the gap is due to an incomplete exploration of the space of the multiplier $\mu$. The example emphasizes a practical problem that often arises when dealing with regularized loss functions: the value of the multiplier has to be thoroughly calibrated by hand, while Moving Targets allows to directly define the desired constraint threshold and is quite robust to different parameter values.

\paragraph{Generalization}
Since our main contribution is an optimization algorithm, we have focused so far on evaluating its performance on the training data, as it simplifies its analysis. We now assess its performance on the test data. In addition to the models of the previous paragraphs, we consider a Random Forest with very low bias (100 estimators with no depth limit), denoted as LBRF, simply trained over the \emph{ideal case} results. Due to the low bias, even this simpler training method obtains feasibility and matches closely the accuracy of the ideal case on the training set.

The results of this evaluation are reported in \Cref{tab:model_generalization}, in the form of average ratio between the scores and the level of constraint satisfaction in the test and the train data. With a few exceptions (e.g. satisfiability in \emph{iris}), the models generalize well in terms of both accuracy and constraint satisfaction. Given the tightness of some of the original constraint and the degree to which the target were altered, this is a remarkable result.
The simpler LBRF approach performs poorly on the test set: while the low bias simplifies training, the price to pay in terms of lack of generalization is quite steep.

\begin{table}[tb]
\centering
\begin{tabular}{lr|cccc}  
\toprule
& & NN & Ens. Trees & LR & LBRF \\
\midrule
Iris 
& $S_{ts} / S_{tr}$
& $ 0.96 $ & $ 0.96 $ & $0.99$ & $0.96$\\
& $C_{ts} / C_{tr}$
& $ 5.68 $ & $ 5.17 $ & $4.31$ & $5.16$\\
Redwine
& $S_{ts} / S_{tr}$
& $0.62$ & $0.92$ & $0.94$ & $0.72$\\
& $C_{ts} / C_{tr}$
& $1.22$ & $1.04$ & $1.35$ & $2.68$\\
Whitewine 
& $S_{ts} / S_{tr}$
& $0.70$ & $0.96$ & $1.00$ & $0.71$\\
& $C_{ts} / C_{tr}$
& $1.11$ & $1.00$ & $0.99$ & $2.92$\\
Shuttle
& $S_{ts} / S_{tr}$
& $0.99$ & $1.00$ & $0.99$ & $1.02$\\
& $C_{ts} / C_{tr}$
& $0.97$ & $1.00$ & $1.01$ & $1.35$\\
Dota2 
& $S_{ts} / S_{tr}$
& $0.83$ & $1.00$ & $0.99$ & $0.58$\\
& $C_{ts} / C_{tr}$
& $1.10$ & $1.00$ & $1.03$ & $2.79$\\
Adult
& $S_{ts} / S_{tr}$
& $0.99$ & $1.00$ & $1.00$ & $0.86$\\
& $C_{ts} / C_{tr}$
& $1.55$ & $1.92$ & $0.98$ & $4.21$\\
Crime
& $S_{ts} / S_{tr}$
& $0.75$ & $0.73$ & $0.93$ & $0.50$\\
& $C_{ts} / C_{tr}$
& $0.74$ & $1.05$ & $1.03$ & $1.53$\\
\bottomrule
\end{tabular}
\caption{Generalization of various models in the test scenario}
\label{tab:model_generalization}
\end{table}

\section{Conclusion}%
\label{sec:Conclusion}

In this paper we have introduced \nmet, a decomposition approach to augment a generic supervised learning algorithm with constraints, by iteratively adjusting the example labels. The method is designed to prioritize constraint satisfaction over accuracy, and proved to behave well on a selection of tasks, constraints, and datasets. Its relative simplicity, reasonable scalability, and the ability to handle any classical ML model and any state-of-the-art constraint solver make it well suited for use in real world settings.

Many open questions remain: we highlighted limitations of regularization based techniques that deserve a much deeper analysis. The convergence properties of our method still need to be characterized. The method scalability should be tested on larger datasets (for which using approximate master steps will be necessary), so as to assess the effect of using meta-heuristics or matheuristics. Given the good performance of the pre-processing approach in some cases \Cref{tab:benchmarks}, it may be interesting to skip the pretraining step in our method. Moreover, since since we allow the use of any ML model, it may be interesting to \emph{combine} \nmet with other approaches for constraint injection in ML.

\section*{Acknowledgment}
This research has been partially funded by the H2020 Project AI4EU,
grant agreement 825619.

\bibliography{aaai2021}

\end{document}